\pdfoutput=1
\documentclass[10pt]{article}

\usepackage[preprint]{tmlr}     %
\usepackage{microtype}
\usepackage{graphicx}
\usepackage{amsmath, amssymb}
\usepackage{booktabs}
\usepackage{xcolor}
\usepackage[colorlinks=true, linkcolor=blue, citecolor=blue, urlcolor=blue]{hyperref}
\makeatletter
\if@preprint
\hypersetup{pdfauthor={Nilotpal Kakati},
            pdftitle={Shared Is Not Enough: Predictor Width and Factor Readability Under Masked Latent Prediction}}
\newcommand{\ifpreprintbuild}[1]{#1}
\newcommand{\availabilitytext}{The code, the simulator, run configurations, and the data behind every figure and table (CSV) are on GitHub at \url{https://github.com/nilotpal09/shared-is-not-enough}; the encoder checkpoints are on Zenodo at \url{https://doi.org/10.5281/zenodo.23559362}. Training is not bitwise reproducible (Appendix~\ref{app:encoder}), so the checkpoints are part of the release.}
\else
\hypersetup{pdfauthor={}, pdftitle={Under review at TMLR}}
\newcommand{\ifpreprintbuild}[1]{}
\newcommand{\availabilitytext}{The code, the simulator, run configurations, and the data behind every figure and table (CSV) are available to reviewers through an anonymised repository, and the encoder checkpoints through an anonymised archive. Training is not bitwise reproducible (Appendix~\ref{app:encoder}), so the checkpoints are part of the release.}
\fi
\makeatother



\newcommand{\simclr}{SimCLR$^{*}$}
\newcommand{\byol}{BYOL$^{*}$}
\newcommand{\mae}{MAE$^{*}$}
\newcommand{\ijepa}{I-JEPA$^{*}$}

\newcommand{\budget}{40{,}000}                 %
\newcommand{\untrainedrefforty}{0.310}         %
\newcommand{\trainedbeststepmax}{5{,}400}      %
\newcommand{\rawinputref}{0.322}  %
\newcommand{\untrainedbeststeprange}{11{,}700 and 39{,}300} %
\newcommand{\probebig}{85k}
\newcommand{\probejointparams}{85{,}189}   %
\newcommand{\probesmallparams}{7{,}811}    %

\newcommand{\pwdefault}{48}
\newcommand{\pwladder}{\{12,\,24,\,48,\,96,\,192\}}   %

\newcommand{\spotdefault}{1000} %
\newcommand{\pddefault}{3} %
\newcommand{\shapereadoutcorr}{0.96} %
\newcommand{\probedrawswords}{two} %

\newcommand{\singleviewmasked}{0.029}    %

\newcommand{\readoutboundcontent}{0.0110}   %
\newcommand{\readoutboundmovement}{0.00208} %

\newcommand{\widthcontentmin}{0.022}       %
\newcommand{\widthcontentmax}{0.229}       %

\newcommand{\floorlevelmed}{0.082}       %

\newcommand{\readoutobjectivegap}{0.10} %

\newcommand{\capacitymovementreference}{0.230} %

\newcommand{\trainingEncoderSpec}{ViT-Nano: 6 layers, width 96, about $0.70$M parameters} %
\newcommand{\trainingPatchSpec}{$4\times4$} %
\newcommand{\trainingOptimiserSpec}{AdamW, batch size 128, bfloat16} %
\newcommand{\trainingLearningRates}{$2\times10^{-4}$ initially, $10^{-3}$ at peak, $10^{-6}$ finally} %
\newcommand{\trainingWeightDecay}{Cosine schedule, $0.04$ to $0.4$} %
\newcommand{\trainingMomentum}{$0.996$ to $1.0$} %
\newcommand{\trainingWarmupSpec}{40-epoch warmup, then cosine decay} %
\newcommand{\trainingEpochSpec}{500 updates over 64{,}000 events} %
\newcommand{\trainingBudgetSpec}{500 epochs} %
\newcommand{\trainingContextMask}{One block covering $[0.85,1.0]$ of the image} %
\newcommand{\trainingTargetMasks}{Four blocks covering $[0.15,0.2]$ each} %
\newcommand{\trainingMinPatches}{10} %
\newcommand{\trainingReconstructionMask}{$0.75$} %
\newcommand{\trainingReconstructionDecoder}{Depth 4, width 48, 3 heads} %
\newcommand{\probeFittingEvents}{10{,}000} %
\newcommand{\probeHeldoutEvents}{1{,}000} %
\newcommand{\probeTestEvents}{1{,}000} %
\newcommand{\referenceLongInits}{8} %

\newcommand{\simVolume}{10\times10\times20\,\mathrm{cm}^3} %
\newcommand{\simGrid}{4\times64\times64} %
\newcommand{\simParticleCount}{5--20} %
\newcommand{\simTotalEnergy}{[100,500]\,\mathrm{GeV}} %
\newcommand{\simFragmentation}{\alpha_{\mathrm{frag}}=5} %
\newcommand{\simPositionWidth}{\sigma_{\mathrm{face}}=0.7\,\mathrm{cm}} %

\newcommand{\energytwinfraction}{81\%} %
\newcommand{\matchedcontrastivefractionwords}{40\%} %
\newcommand{\constantmedaezone}{0.665} %

\newcommand{\constantshapealpha}{4.456}
\newcommand{\constantshapeT}{6.606}

\title{Shared Is Not Enough: Predictor Width and Factor Readability Under Masked Latent Prediction}

\author{\name Nilotpal Kakati \email nilotpal.kakati09@gmail.com \\
        \addr Weizmann Institute of Science, Rehovot, Israel}

\begin{document}
\maketitle

\begin{abstract}
Masked latent prediction methods such as I-JEPA aim to learn predictable
features, but what determines which information they retain is less clear.
We built a simulator in which every factor behind an image is known, so a
probe can measure how readable each factor remains after training. We trained small encoders (0.7M parameters) with I-JEPA, SimCLR and BYOL on the same pairs of views and probed a
factor that both views share. I-JEPA had a predictor with half the encoder's
width and depth, the proportions it uses for ViT-B. Its median score was about
\matchedcontrastivefractionwords{} of SimCLR's.
Widening the predictor to the encoder's width reversed this ordering at every
training seed, against both other methods.
We release the simulator and the trained models.

\end{abstract}

\section{Introduction}
\label{sec:intro}
Masked latent prediction trains an encoder by predicting the representations of hidden image regions from the visible ones. Its designers argue that predicting in representation space favours information the predictor can recover and may leave out unpredictable details \citep{lecun2022path,assran2023ijepa,garrido2024iwm}. Downstream performance shows how useful the representations are for a task, but on its own does not show which input factors stay readable after training.

Prior work has asked whether known factors can be recovered from learned representations, in theory and in controlled experiments \citep{vonkugelgen2021content,pendharkar2026predictiveobjectives}. When augmentations change some information between views, the predictor's capacity affects how much of it the encoder keeps \citep{garrido2024iwm}. We ask whether a factor that both views share stays readable under masked latent prediction, and how that depends on the predictor.

\begin{figure}[!htb]
  \centering
  \includegraphics[width=\textwidth,trim=0 6.2bp 0 2.4bp,clip]{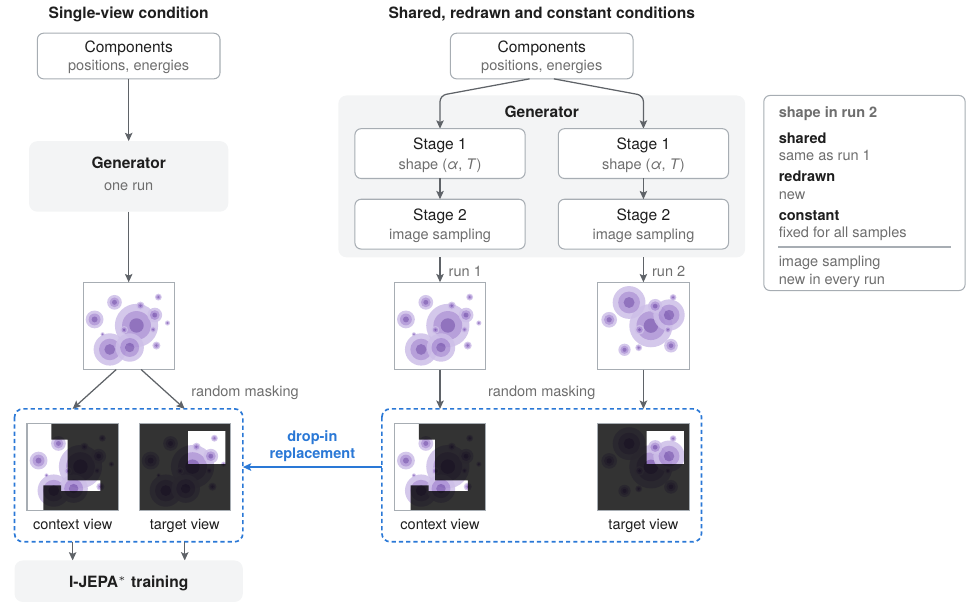}
  \caption{The view-pair construction. The single-view condition (left) cuts
  both views from one image. The other conditions (right) run the simulator
  twice on the same components: run 1 gives the context view and run 2 the
  target view. Each run draws shape (stage 1) and samples the image (stage 2)
  with separate seeds. Run 2 reuses run 1's shape in the shared condition and
  draws a new one in the redrawn condition. In the constant condition, every
  component of every image has the same shape. Masking, predictor and target
  encoder are unchanged; only the source of the target view changes.}
  \label{fig:pair}
\end{figure}

A factor that both views share is not always easy to recover when each image shows it only through noise. In scientific imaging, such a factor can still be the quantity of interest. Our simulator produces such images with every factor's true value known. It keeps the factors behind each image separate from the random sampling that draws it, and the number of sampled points sets how grainy the image is. This separation lets a factor's value be shared across the two views, redrawn between them, or held constant over all training images (Figure~\ref{fig:pair}). In all three cases, the two views are sampled separately, so they do not share the noise.

We compared masked latent prediction (\ijepa{}) with \simclr{} and \byol{}, all trained on the same views with the same encoder architecture (Section~\ref{sec:apparatus}). The star marks our implementations. The default predictor is shallower and narrower than the encoder. We also matched it to the encoder in depth or in width, with all other settings unchanged. A probe fitted to a factor's known values measures how readable that factor is from the frozen representations.

Our contributions are:
\begin{itemize}
\item With the default predictor, masked latent prediction left a shared factor less readable than \simclr{} and \byol{} did. At the end of training, its median score was about \matchedcontrastivefractionwords{} of \simclr{}'s. Holding the factor constant gave similarly low scores, and redrawing it between views raised the median.
\item Making the predictor deeper or wider raised readability at every training seed. Making it as wide as the encoder reversed the ordering against both at every seed.
\item We release the simulator and the trained models. Because every factor's true value is known, other objectives and predictors can be tested the same way.
\end{itemize}

\section{Setup}
\label{sec:apparatus}

\subsection{Simulator}
\begin{figure}[t]
  \centering
  \includegraphics[width=\textwidth,trim=0 9.4bp 0 1.5bp,clip]{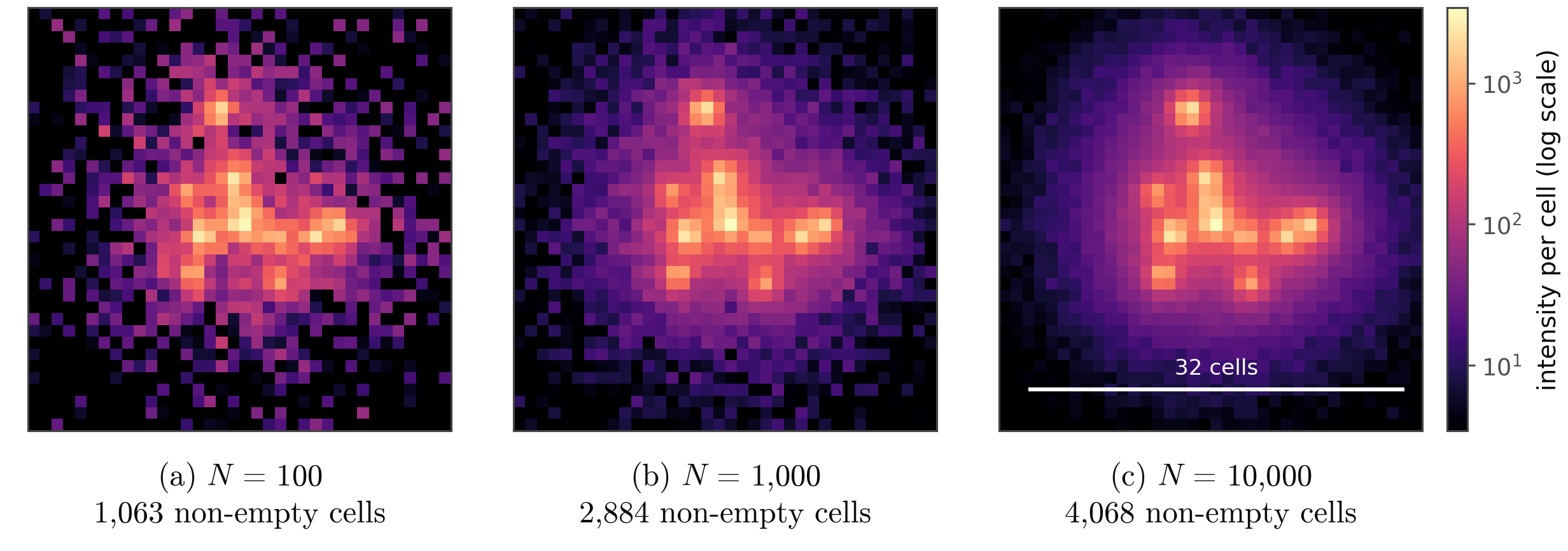}
  \caption{One image at three graininess settings $N$, summed over the four
  channels and cropped to a $36\times36$ window around the components.
  Component positions, energies and shapes are held fixed, so only
  the image sampling differs between panels; the number of non-empty cells in
  the full $64\times64$ grid is given beneath each panel.}
  \label{fig:nspots}
\end{figure}

Each image has four channels and shows several overlapping patterns, which
we call components (Figure~\ref{fig:nspots}). Each component has a known
position, energy $E$ and shape parameters $(\alpha,T)$. The shape parameters
set how the component's intensity is split across the four channels and how
far it spreads within each. Shape is the factor this paper studies.
Appendix~\ref{app:sim} describes the physical model behind the simulator.

The simulator first draws each component's shape from a distribution that
depends on its energy. It then samples points from the
resulting spatial patterns and bins their contributions onto the image. The
number of points per component per channel, $N$, controls image graininess.
Larger $N$ gives smoother images (Figure~\ref{fig:nspots}). The two stages use separate random streams,
so we can redraw the shape or the image sampling separately,
keeping component positions and energies fixed. Training and probe images use
$N=\spotdefault{}$ unless stated otherwise.

\subsection{Training Conditions and Objectives}
Each training pair consists of two views: images that show the same components, at the same positions and energies, but are sampled separately. In the \emph{shared} condition, each component keeps the same shape in both views, so shape is the factor they share. The main comparison uses this condition: \ijepa{} with three predictor sizes against \simclr{} and \byol{}. Three further conditions support the secondary results (Table~\ref{tab:conditions}). In the \emph{redrawn} condition, shape is drawn separately for each view. In the \emph{constant} condition, every component of every training image has the same shape, whatever its energy (Appendix~\ref{app:sim}). The \emph{single-view} condition instead takes both views from one generated image (Figure~\ref{fig:pair}).

\begin{table}[t]
\centering
\caption{Training conditions and the objectives trained on each. The shared condition carries the main comparison. No redrawn \byol{} was trained.}
\label{tab:conditions}
\resizebox{\textwidth}{!}{%
\begin{tabular}{lll}
\toprule
Condition & Shape across the training images & Objectives \\
\midrule
Shared & Same in both views; varies across pairs & \ijepa{} (three predictors), \simclr{}, \byol{} \\
\midrule
Redrawn & Drawn separately for each view & \ijepa{}, \simclr{} \\
Constant & Fixed for all training images & \ijepa{}, \simclr{}, \byol{} \\
Single-view & One image for both views; varies across images & \ijepa{}, \mae{} \\
\bottomrule
\end{tabular}}
\end{table}

\textbf{Encoder and objectives.} A vision transformer \citep{dosovitskiy2021vit} maps each spatial patch, across all four channels, to a token. Masking hides rectangular regions of this patch grid. Each hidden patch loses all its channels. \ijepa{}, our implementation of masked latent prediction, uses the visible context to predict the target encoder's representations of the hidden patches \citep{assran2023ijepa}. Visible and hidden regions can both carry evidence about a component's shape.

We compared \ijepa{} with \simclr{} \citep{chen2020simclr} and \byol{} \citep{grill2020byol}, trained with the same encoder architecture, data, optimiser and training budget. Both encode the two views without masking and mean-pool their tokens before projection (Appendix~\ref{app:training}). Masking, pooling, prediction heads and target construction differ between objectives. Comparisons between \ijepa{} predictors keep masking, pooling and target construction fixed. \mae{} \citep{he2022mae}, which reconstructs the hidden pixels, is an additional comparison on the single-view images and a positive control for the probe (Appendix~\ref{app:training}). All objectives train on views from the simulator with no further image augmentations.

The default \ijepa{} predictor has width \pwdefault{} and depth \pddefault{}, and we varied both. Appendix~\ref{app:training} gives head sizes, architectures and optimisation settings.

Encoders are named by condition and objective, as in shared \ijepa{}, adding the predictor or training graininess when either differs from the default. Each setting in the main figures has one encoder for each of six training seeds. The main comparisons share their seeds, so their encoders can be compared seed by seed.

\subsection{Readability}\label{sec:readability}

A probe predicts component parameters from the tokens of a frozen encoder. It receives a query at each component's known position and attends across the unpooled token sequence. We fitted the probe on separately generated images, selected its checkpoint on one held-out set and scored it on another. Each encoder received \probedrawswords{} independent fits, and main-text figures average their scores.

Its targets are energy and two shape coordinates, $z_1$ and $z_2$, predicted jointly. The shape coordinates are built from $(\alpha,T)$ to be independent of energy (Appendix~\ref{app:training}), so a probe cannot score on them by reading energy alone. The scores of $z_1$ and $z_2$ track each other across settings, so we report $z_1$ as the readability of shape.
For each target, the score is the improvement in median
absolute error over a baseline that always predicts the median of the scored set:

\[
d = \operatorname{MedAE}_{\mathrm{baseline}} - \operatorname{MedAE}_{\mathrm{probe}}.
\]

Higher scores mean better readability, and $d(z_1)$ is the score on $z_1$. On $z_1$, $d=0.1$ means the probe removes about 15\% of the baseline's error.

An untrained encoder, read with the same probe and fitting budget, gives a reference level. On this probe its median score for $z_1$ was above every trained encoder's. Every comparison in this paper is between trained encoders read by the same probe, so none depends on this gap (Section~\ref{sec:outlook}).

For \ijepa{} and \byol{}, the probe reads the target encoder, the moving-average copy. For \ijepa{}, reading the online copy instead changed its scores far less than its gap to \simclr{} (Appendix~\ref{app:training}).

\section{Results}\label{sec:results}
The main result is in Section~\ref{sec:readable}. With the default predictor, shape was less readable from \ijepa{} than from \simclr{} and \byol{}. Making the predictor as wide as the encoder reversed this. Sections~\ref{sec:diet} and~\ref{sec:movement} then ask what in the training data makes shape more readable, and whether a measure that needs no probe ranks the encoders the same way.

\subsection{Readability of the Shared Factor}\label{sec:readable}\label{sec:capacity}

\begin{figure}[!htb]
  \centering
  \includegraphics[width=4.7in,trim=0 19.9bp 0 6.9bp,clip]{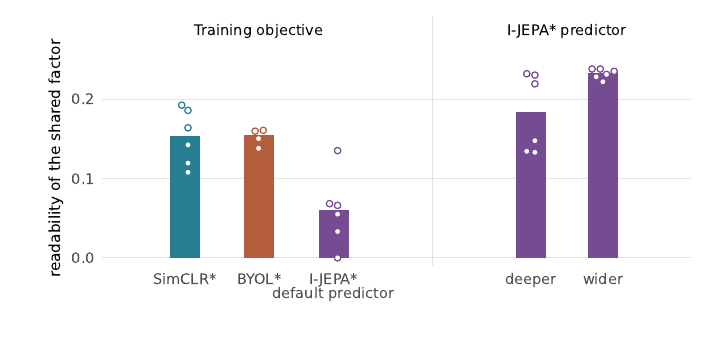}
  \caption{Readability of the shared factor, $d(z_1)$, after training on the same view pairs.
  A probe predicts the known shape coordinate from frozen encoder tokens. Bars show medians;
  circles show training seeds, each averaged over two probe fits. Each setting has six seeds;
  two collapsed \byol{} encoders are excluded (Appendix~\ref{app:training}). An untrained
  encoder's median score on this probe is \untrainedrefforty{} (Section~\ref{sec:readability}).}
  \label{fig:core}
\end{figure}

At the end of training, the median readability of default \ijepa{} was about \matchedcontrastivefractionwords{} of \simclr{}'s (Figure~\ref{fig:core}), and \byol{} scored like \simclr{} once its collapsed encoders were left out (Appendix~\ref{app:training}). Five of the six \ijepa{} encoders scored below every encoder of the other two objectives. Earlier in training the order was reversed: from epoch 20 to 100 of 500, \simclr{} scored below \ijepa{} at every seed (Appendix~\ref{app:seeds}).

The default \ijepa{} predictor has half the encoder's width and depth, the proportions the original I-JEPA uses for ViT-B; its larger released models use predictors narrower still \citep{assran2023ijepa}. We made the predictor as wide or as deep as the encoder, with the same seeds and all other settings unchanged. A larger predictor raised readability at every seed. With the wider predictor, every \ijepa{} encoder scored above every \simclr{} and \byol{} encoder (Figure~\ref{fig:core}). The deeper predictor split the seeds: three reached the wider predictor's level and three stayed within the \simclr{} and \byol{} range. Single-view sweeps of width and depth showed the same rise (Table~\ref{tab:singleviewcapacity}).

Energy is also the same in both views, but the gap there was smaller: shared \ijepa{} reached about \energytwinfraction{} of shared \simclr{}'s energy score, against \matchedcontrastivefractionwords{} for shape. Widening the predictor raised both scores. On energy too, every encoder with the wider predictor scored above every \simclr{} and \byol{} encoder (Appendix~\ref{app:energy}).

Readability depends on the probe, so we refitted the shared encoders with three other probes (Table~\ref{tab:instruments}). On all three, the default predictor's median stayed below \simclr{}'s and \byol{}'s. The wider predictor stayed ahead of both at every seed on two of the probes; on the third, which normalises the encoder's tokens differently, it was above every \byol{} encoder and above \simclr{} at five of the six seeds. The lead also held with narrower projection heads (Appendix~\ref{app:comparison}).

\subsection{What Raises Readability}\label{sec:diet}

\begin{figure}[!htb]
  \centering
  \includegraphics[width=6.0in,trim=0 10.5bp 0 5.5bp,clip]{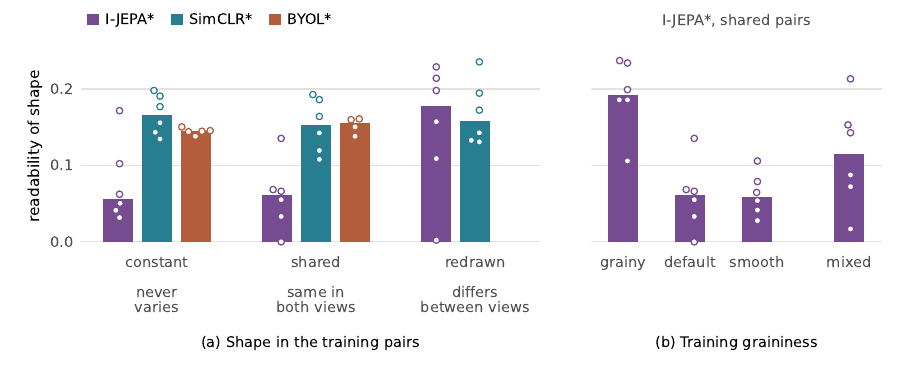}
  \caption{Readability of shape, $d(z_1)$, at the end of training. All \ijepa{} encoders use the default predictor; all probes use the default graininess. The default bar in (b) is the shared \ijepa{} bar in (a). Bars show medians; circles show training seeds, each averaged over two probe fits. Each setting has six seeds; three collapsed \byol{} encoders are excluded, one constant and two shared (Appendix~\ref{app:training}). No redrawn \byol{} was trained.}
  \label{fig:support}
\end{figure}

Redrawing shape between views raised the median readability of default \ijepa{}, and five of the six seeds scored higher than in the shared condition (Figure~\ref{fig:support}a). The sixth scored near zero, and the redrawn scores overlapped those of redrawn \simclr{}.

In the constant condition, shape never varied across training images. \ijepa{} trained this way scored about as low as in the shared condition (Figure~\ref{fig:support}a). So when both views shared shape, seeing it vary across images added little. The \simclr{} and \byol{} medians were about the same in both conditions and stayed ahead. Shape varied freely in all three probe sets, including for encoders trained in the constant condition.

We also varied the graininess of the training images. Besides the default $N=\spotdefault{}$, shared \ijepa{} was trained on \emph{grainy} ($N=100$), \emph{smooth} ($N=10^4$) and \emph{mixed} images, with $N$ sampled log-uniformly from 100 to $10^4$. On default probe images, training on grainy images gave the highest score, at the top of the shared \simclr{} and \byol{} range (Figure~\ref{fig:support}b). Mixed images gave a middle score, and smooth images scored like default ones. This order, grainy first and mixed second, held at all three probe graininesses and in the single-view condition (Appendix~\ref{app:diet}).

\subsection{Movement}\label{sec:movement}

Readability depends on a fitted probe. Movement needs no probe, and we tested whether it orders encoders the same way. It measures how much an image's representation changes when only its shape is redrawn. Each redraw gives a replica of the image, with the same component positions, energies and image-sampling seed; only $(\alpha,T)$ changes. We call this factor-only resampling. Movement is the mean cosine distance between replicas of the same image divided by the mean distance between different images, computed per token (Appendix~\ref{app:training}).

Higher movement means shape changes the representation more. Replicas already look alike at the input, so we compared movement with that of an untrained encoder. Collapsed encoders bring both distances near zero, which makes the ratio unreliable, so we excluded the collapsed \byol{} encoders.

\begin{figure}[!htb]
  \centering
  \includegraphics[width=\textwidth,trim=0 9.1bp 0 5.7bp,clip]{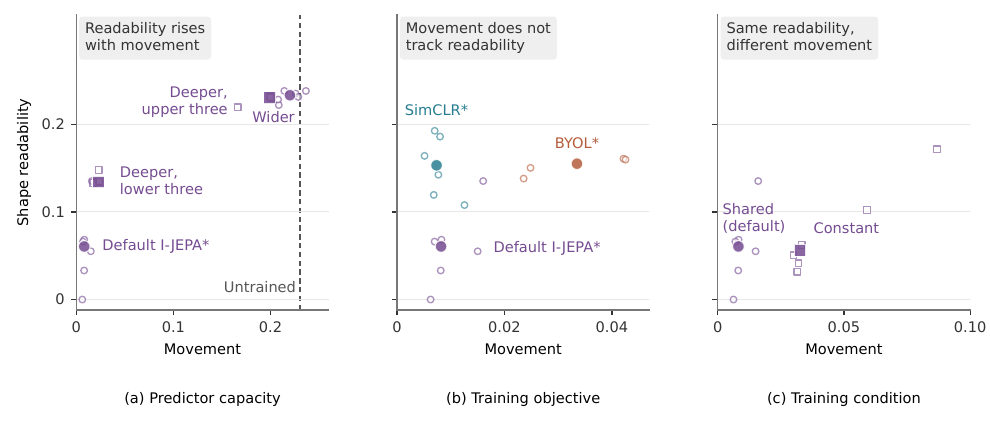}
  \caption{Per-token movement and shape readability, $d(z_1)$, rose together
  across predictor settings (a). \simclr{} had higher readability than default
  \ijepa{} at overlapping movement; \byol{} had similar readability to
  \simclr{} at higher movement (b). Shared and constant \ijepa{} differed in
  movement but overlapped in readability (c). Default shared \ijepa{} appears
  in all panels. Open markers show seeds; filled markers show medians, with
  the deeper predictor's two groups summarised separately. Movement uses
  factor-only resampling, and the dashed line in (a) marks an untrained
  encoder, \capacitymovementreference{}. Horizontal scales differ.}
  \label{fig:dissociation}
\end{figure}

Movement and readability rose together as the predictor grew
(Figure~\ref{fig:dissociation}a). Every wider-predictor seed had movement near
the untrained reference. Within the deeper predictor, the three seeds with
higher readability also moved more.

Across objectives, movement did not track readability
(Figure~\ref{fig:dissociation}b). Shared default \ijepa{} overlapped \simclr{} in movement but had lower median
readability. This held for both its online and target copies (Appendix~\ref{app:training}).
\byol{} moved more than \simclr{},
while their readability ranges overlapped.

Across training conditions, with the default predictor, every constant \ijepa{} encoder moved more than every shared one, yet their readability scores overlapped (Figure~\ref{fig:dissociation}c). So of the three comparisons, movement tracked readability only as the predictor grew. In the other two, it could not stand in for the probe.

\section{Related Work}\label{sec:related}

Prior work examines when shared factors remain readable. \citet{vonkugelgen2021content} give a guarantee for recovering shared content under an invertible observation model, where the factors can in principle be recovered from one image. Controlled experiments show that an easy shared feature can suppress competing features under contrastive losses \citep{chen2021intriguing}. Under predictive learning, known factors have also been probed directly \citep{sobal2022slowfeatures,pendharkar2026predictiveobjectives}. In the closest of these, \citet{pendharkar2026predictiveobjectives} varied a feature's persistence under JEPA-style prediction of future latents, and a feature held fixed across the pair stayed readable. Ours became less readable under \ijepa{} with the default predictor than under \simclr{} and \byol{}. The tasks differ, and so does the way the factor enters the image.

Image World Models (IWM) showed that predictor capacity affects how much information about the change between views the encoder keeps \citep{garrido2024iwm}. In their words, with a weak predictor ``the encoder removes the information that cannot be predicted''. In our experiments, predictor depth and width also mattered for a factor shared across views. Theory offers candidate explanations. In linear models, feature learning depends on statistics across views and on training dynamics \citep{littwin2024jepanoisy,tian2021understanding}. A trainable prediction head can also help weaker features be learned \citep{wen2022predictionhead}.

\section{Discussion and Conclusion}\label{sec:outlook}

Sharing the factor across views did not protect its readability under the default predictor. Widening the predictor put \ijepa{} above \simclr{} and \byol{} on the same data. One hypothesis is that a small predictor favours encoder features that are easier to predict, at the expense of a weakly expressed factor. IWM make a similar argument for information that changes between views \citep{garrido2024iwm}. With the default predictor, \ijepa{} reached a larger fraction of the \simclr{} score for energy than for shape, and widening raised both scores (Appendix~\ref{app:energy}). I-JEPA's ablations favour a narrower predictor for downstream evaluation \citep[Tables~12 and~14]{assran2023ijepa}, whereas matching predictor and encoder width improved this factor's readability.

Readability also depended on the training data. Redrawing shape between views raised the score, and so did training on grainier images.

Movement rose with readability when the predictor grew, but it did not consistently order readability across objectives or training conditions (Section~\ref{sec:movement}). These experiments cover one factor in one simulator, with small encoders. We have not established why the predictor or the training data changed readability.

Every trained encoder scored below the untrained median (Section~\ref{sec:apparatus}). Replacing the probe's token layer norm with fixed standardisation did not close this gap for the encoders in Table~\ref{tab:instruments} or for \mae{} (Appendix~\ref{app:training}). We cannot tell whether training removed the shape information or only made it harder to read. Our comparisons are between trained encoders read by the same probe, so they do not depend on the answer.

A direct next test is to hold a measurable factor shared across natural-image views and vary predictor capacity with the encoder and training schedule fixed. We expect a weakly expressed shared factor to be less readable under I-JEPA's default predictor proportions than under \simclr{} on the same views. Increasing predictor capacity should narrow or reverse that gap.

\section*{Code and Data Availability}
\availabilitytext

\section*{LLM Usage}
Large language models assisted with coding, experiment execution, analysis,
figure production, and manuscript drafting and editing under the author's
direction and review. Study design, verification of the reported results, and
scientific interpretation remained with the author, who takes responsibility
for the work.

\ifpreprintbuild{\section*{Acknowledgements}
The author thanks Eilam Gross for discussions and feedback on the manuscript.
This work was supported by the Minerva Grant No. 7015027, the Knell Family
Institute for Artificial Intelligence, the Benoziyo Center for High Energy
Physics, a BSF Grant No. 2024220, and the Krenter-Perinot Center for High
Energy Particle Physics.}

\bibliography{references}

\begin{thebibliography}{18}
\providecommand{\natexlab}[1]{#1}
\providecommand{\url}[1]{\texttt{#1}}
\expandafter\ifx\csname urlstyle\endcsname\relax
  \providecommand{\doi}[1]{doi: #1}\else
  \providecommand{\doi}{doi: \begingroup \urlstyle{rm}\Url}\fi

\bibitem[Assran et~al.(2023)Assran, Duval, Misra, Bojanowski, Vincent, Rabbat,
  LeCun, and Ballas]{assran2023ijepa}
Mahmoud Assran, Quentin Duval, Ishan Misra, Piotr Bojanowski, Pascal Vincent,
  Michael Rabbat, Yann LeCun, and Nicolas Ballas.
\newblock Self-supervised learning from images with a joint-embedding
  predictive architecture.
\newblock In \emph{Proceedings of the IEEE/CVF Conference on Computer Vision
  and Pattern Recognition (CVPR)}, pp.\  15619--15629, 2023.
\newblock \doi{10.1109/CVPR52729.2023.01499}.

\bibitem[Chen et~al.(2020)Chen, Kornblith, Norouzi, and Hinton]{chen2020simclr}
Ting Chen, Simon Kornblith, Mohammad Norouzi, and Geoffrey Hinton.
\newblock A simple framework for contrastive learning of visual
  representations.
\newblock In \emph{Proceedings of the 37th International Conference on Machine
  Learning (ICML)}, volume 119 of \emph{Proceedings of Machine Learning
  Research}, pp.\  1597--1607, 2020.

\bibitem[Chen et~al.(2021)Chen, Luo, and Li]{chen2021intriguing}
Ting Chen, Calvin Luo, and Lala Li.
\newblock Intriguing properties of contrastive losses.
\newblock In \emph{Advances in Neural Information Processing Systems 34
  (NeurIPS 2021)}, 2021.
\newblock URL
  \url{https://proceedings.neurips.cc/paper_files/paper/2021/hash/628f16b29939d1b060af49f66ae0f7f8-Abstract.html}.

\bibitem[Dosovitskiy et~al.(2021)Dosovitskiy, Beyer, Kolesnikov, Weissenborn,
  Zhai, Unterthiner, Dehghani, Minderer, Heigold, Gelly, Uszkoreit, and
  Houlsby]{dosovitskiy2021vit}
Alexey Dosovitskiy, Lucas Beyer, Alexander Kolesnikov, Dirk Weissenborn,
  Xiaohua Zhai, Thomas Unterthiner, Mostafa Dehghani, Matthias Minderer, Georg
  Heigold, Sylvain Gelly, Jakob Uszkoreit, and Neil Houlsby.
\newblock An image is worth 16x16 words: Transformers for image recognition at
  scale.
\newblock In \emph{International Conference on Learning Representations
  (ICLR)}, 2021.

\bibitem[Dreyer et~al.(2026)Dreyer, Kakati, and Gross]{dreyer_pycaloflash}
Etienne Dreyer, Nilotpal Kakati, and Eilam Gross.
\newblock {PyCaloFlash}: {GPU}-ready fast {EM} shower model.
\newblock Zenodo, version v1.0, 2026.
\newblock URL \url{https://doi.org/10.5281/zenodo.20588830}.

\bibitem[Garrido et~al.(2024)Garrido, Assran, Ballas, Bardes, Najman, and
  LeCun]{garrido2024iwm}
Quentin Garrido, Mahmoud Assran, Nicolas Ballas, Adrien Bardes, Laurent Najman,
  and Yann LeCun.
\newblock Learning and leveraging world models in visual representation
  learning.
\newblock \emph{arXiv preprint arXiv:2403.00504}, 2024.

\bibitem[Grill et~al.(2020)Grill, Strub, Altch{\'e}, Tallec, Richemond,
  Buchatskaya, Doersch, Avila~Pires, Guo, Gheshlaghi~Azar, Piot, Kavukcuoglu,
  Munos, and Valko]{grill2020byol}
Jean-Bastien Grill, Florian Strub, Florent Altch{\'e}, Corentin Tallec,
  Pierre~H. Richemond, Elena Buchatskaya, Carl Doersch, Bernardo Avila~Pires,
  Zhaohan~Daniel Guo, Mohammad Gheshlaghi~Azar, Bilal Piot, Koray Kavukcuoglu,
  R{\'e}mi Munos, and Michal Valko.
\newblock Bootstrap your own latent: A new approach to self-supervised
  learning.
\newblock In \emph{Advances in Neural Information Processing Systems
  (NeurIPS)}, 2020.

\bibitem[Grindhammer \& Peters(2000)Grindhammer and
  Peters]{grindhammer2000gflash}
G.~Grindhammer and S.~Peters.
\newblock The parameterized simulation of electromagnetic showers in
  homogeneous and sampling calorimeters.
\newblock \emph{arXiv preprint hep-ex/0001020}, 2000.

\bibitem[Harris et~al.(2025)Harris, Krupa, Kagan, Maier, and
  Woodward]{harris2025rs3l}
Philip Harris, Jeffrey Krupa, Michael Kagan, Benedikt Maier, and Nathaniel
  Woodward.
\newblock Resimulation-based self-supervised learning for pretraining physics
  foundation models.
\newblock \emph{Physical Review D}, 111:\penalty0 032010, 2025.

\bibitem[He et~al.(2022)He, Chen, Xie, Li, Doll{\'a}r, and Girshick]{he2022mae}
Kaiming He, Xinlei Chen, Saining Xie, Yanghao Li, Piotr Doll{\'a}r, and Ross
  Girshick.
\newblock Masked autoencoders are scalable vision learners.
\newblock In \emph{Proceedings of the IEEE/CVF Conference on Computer Vision
  and Pattern Recognition (CVPR)}, 2022.

\bibitem[LeCun(2022)]{lecun2022path}
Yann LeCun.
\newblock A path towards autonomous machine intelligence, 2022.
\newblock URL \url{https://openreview.net/forum?id=BZ5a1r-kVsf}.
\newblock Version 0.9.2, 2022-06-27. OpenReview.

\bibitem[Littwin et~al.(2024)Littwin, Saremi, Advani, Thilak, Nakkiran, Huang,
  and Susskind]{littwin2024jepanoisy}
Etai Littwin, Omid Saremi, Madhu Advani, Vimal Thilak, Preetum Nakkiran, Chen
  Huang, and Joshua Susskind.
\newblock How {JEPA} avoids noisy features: The implicit bias of deep linear
  self distillation networks.
\newblock In \emph{Advances in Neural Information Processing Systems 37
  (NeurIPS 2024)}, 2024.
\newblock URL
  \url{https://proceedings.neurips.cc/paper_files/paper/2024/hash/a600f0f740605205133553cb74a1c107-Abstract-Conference.html}.

\bibitem[Pendharkar(2026)]{pendharkar2026predictiveobjectives}
Ayan Pendharkar.
\newblock Predictive objectives discard exogenous control-relevant features: A
  controlled mechanistic study, 2026.
\newblock arXiv preprint, not peer reviewed.

\bibitem[Rieck et~al.(2025)Rieck, Cranmer, Dreyer, Gross, Kakati, Kobylianskii,
  Merz, and Soybelman]{rieck2025ssljet}
Patrick Rieck, Kyle Cranmer, Etienne Dreyer, Eilam Gross, Nilotpal Kakati,
  Dmitrii Kobylianskii, Garrett~W. Merz, and Nathalie Soybelman.
\newblock Self-supervised learning strategies for jet physics.
\newblock \emph{Machine Learning: Science and Technology}, 6\penalty0
  (4):\penalty0 045015, 2025.
\newblock \doi{10.1088/2632-2153/ae1100}.

\bibitem[Sobal et~al.(2022)Sobal, {S V}, Jalagam, Carion, Cho, and
  LeCun]{sobal2022slowfeatures}
Vlad Sobal, Jyothir {S V}, Siddhartha Jalagam, Nicolas Carion, Kyunghyun Cho,
  and Yann LeCun.
\newblock Joint embedding predictive architectures focus on slow features,
  2022.
\newblock Short paper, NeurIPS 2022 Workshop on Self-Supervised Learning:
  Theory and Practice.

\bibitem[Tian et~al.(2021)Tian, Chen, and Ganguli]{tian2021understanding}
Yuandong Tian, Xinlei Chen, and Surya Ganguli.
\newblock Understanding self-supervised learning dynamics without contrastive
  pairs.
\newblock In \emph{Proceedings of the 38th International Conference on Machine
  Learning (ICML)}, volume 139 of \emph{Proceedings of Machine Learning
  Research}, pp.\  10268--10278, 2021.

\bibitem[von K{\"u}gelgen et~al.(2021)von K{\"u}gelgen, Sharma, Gresele,
  Brendel, Sch{\"o}lkopf, Besserve, and Locatello]{vonkugelgen2021content}
Julius von K{\"u}gelgen, Yash Sharma, Luigi Gresele, Wieland Brendel, Bernhard
  Sch{\"o}lkopf, Michel Besserve, and Francesco Locatello.
\newblock Self-supervised learning with data augmentations provably isolates
  content from style.
\newblock In \emph{Advances in Neural Information Processing Systems 34
  (NeurIPS 2021)}, 2021.
\newblock URL
  \url{https://proceedings.neurips.cc/paper_files/paper/2021/hash/8929c70f8d710e412d38da624b21c3c8-Abstract.html}.

\bibitem[Wen \& Li(2022)Wen and Li]{wen2022predictionhead}
Zixin Wen and Yuanzhi Li.
\newblock The mechanism of prediction head in non-contrastive self-supervised
  learning.
\newblock In \emph{Advances in Neural Information Processing Systems 35
  (NeurIPS 2022)}, 2022.

\end{thebibliography}
\bibliographystyle{tmlr}

\clearpage
\appendix
\section{Simulator Details}
\label{app:sim}
The components in Section~\ref{sec:apparatus} are simulated particles. Each
particle enters a block of lead at a specified position and deposits its
energy over a spreading pattern, called a shower. The simulator records this
pattern on a grid, with depth layers serving as image channels. We used
PyCaloFlash v1.0 \citep{dreyer_pycaloflash}, a PyTorch implementation of
the parameterised shower model of \citet{grindhammer2000gflash}, with its two
generation stages drawn from independent random streams.

An image combines the deposits from several particles, whose showers are
generated separately and added on the same grid. Particle energies sum to the
image's total energy, and their entry positions are spread around a common
axis (Table~\ref{tab:simulatorconfig}).

\begin{table}[ht]
\centering
\caption{Geometry and particle sampling used to generate the images. The
Dirichlet distribution sets the particles' shares of the total energy.}
\label{tab:simulatorconfig}
\begin{tabular}{ll}
\toprule
Setting & Value \\
\midrule
Material and volume & Lead, $\simVolume{}$ \\
Grid (depth $\times$ spatial dimensions) & $\simGrid{}$ \\
Particles per image & \simParticleCount{} \\
Total energy & $\simTotalEnergy{}$ \\
Energy fractions & Dirichlet, $\simFragmentation{}$ \\
Entry positions about the axis & Gaussian, $\simPositionWidth{}$ \\
\bottomrule
\end{tabular}
\end{table}

\textbf{Shape generation.} Stage 1 draws $(\alpha,T)$ separately for each
particle, from a distribution whose means, widths and correlation depend on
the particle's energy. The pair defines a gamma density for the energy
deposited along the depth direction: $T$ locates its maximum and $\alpha$
controls its shape.

The shape parameters also set the spread within each layer: the radial
kernel depends on depth divided by the profile's mean depth, a quantity
called shower age (equations 24--26, 28 and 34 of
\citet{grindhammer2000gflash}, as implemented).

In the constant condition every component has $\alpha=\constantshapealpha{}$ and
$T=\constantshapeT{}$, independent of its energy.

\textbf{Image sampling.} Stage 2 turns the profile into discrete deposits.
The depth profile sets each layer's total energy, and the radial kernel
places $N$ points within each layer, which are then binned onto the grid.
More points reduce the fluctuations between independent
samplings of the same profile (Figure~\ref{fig:nspots}). Re-simulating the
same particles gives independent images, as in the paired views of
Section~\ref{sec:apparatus} and in earlier studies that re-ran the
simulation \citep{harris2025rs3l, rieck2025ssljet}.

\section{Training and Evaluation Details}
\label{app:training}

\subsection{Encoder Training}\label{app:encoder}

The main experiments used the I-JEPA optimisation recipe
\citep{assran2023ijepa}, with the settings in
Table~\ref{tab:trainingconfig}.

A token embeds a $4\times4$ tile of the $4\times64\times64$ image across all
four channels, giving a $16\times16$ grid of 256 tokens. Cell energies, in MeV, are transformed as
\[
  x_{\mathrm{cell}}=\frac{\log(1+E_{\mathrm{cell}})}{\log(1+5500)}.
\]
Empty cells map to zero, and occupied cells take values up to about 1.1.

\begin{table}[t]
\centering
\caption{Default encoder-training settings. Data are simulated on the fly.}
\label{tab:trainingconfig}
\begin{tabular}{ll}
\toprule
Setting & Value \\
\midrule
Encoder & \trainingEncoderSpec{} \\
Patch size & \trainingPatchSpec{} \\
\ijepa{} predictor & Width \pwdefault{}, depth \pddefault{} \\
Optimiser & \trainingOptimiserSpec{} \\
Learning rate & \trainingLearningRates{} \\
Weight decay & \trainingWeightDecay{} \\
Target EMA momentum & \trainingMomentum{} \\
Learning-rate schedule & \trainingWarmupSpec{} \\
Epoch size & \trainingEpochSpec{} \\
Training budget & \trainingBudgetSpec{} \\
Context mask & \trainingContextMask{} \\
Target masks & \trainingTargetMasks{} \\
Minimum patches kept & \trainingMinPatches{} \\
\mae{} mask ratio & \trainingReconstructionMask{} \\
\mae{} decoder & \trainingReconstructionDecoder{} \\
\bottomrule
\end{tabular}
\end{table}

The warmup corresponds to about 20{,}000 updates. The predictor sweeps kept
ViT-Nano fixed and varied width over $\pwladder{}$ or depth over $\{1,3,6\}$.
The shared comparison used predictors of width/depth 48/3 (default), 48/6
(deeper) and 96/3 (wider), all at epoch 500. I-JEPA's released ViT-H and
ViT-g checkpoints use predictor-to-encoder width ratios of 0.30 and 0.27
\citep{assran2023ijepa}.

Training is not bitwise reproducible with the default GPU
kernels: two same-seed runs diverged within five epochs. With deterministic
kernels, two same-seed runs gave identical checkpoints.

\subsection{Comparison Settings}\label{app:comparison}

\simclr{} and \byol{} used the same configurations as \ijepa{} outside the
objective block. Both encode the two views without masking and mean-pool
their tokens before projection. Each projector is a two-layer MLP with
dimensions $96 \to 384 \to 96$, a GELU activation and no batch
normalisation. \byol{} adds a predictor with the same dimensions; its target
encoder and projector are exponential moving averages of the online modules.
\simclr{} applies InfoNCE to unit-normalised projections with temperature
0.2. \byol{} minimises the squared distance between unit-normalised
predictions and the other view's target projections, averaged over both
directions. These heads are discarded for probing.

Narrowing the hidden width of both projectors from 384 to 96 or 48, with the
\byol{} predictor kept at 384, did not lower either method's median score in
the shared condition. \byol{} collapsed in two, three and one of six seeds at
widths 384, 96 and 48. At every projector width, every shared \ijepa{} encoder
with the wider predictor still scored above every non-collapsed encoder of
either method.

With the default projector, the \byol{} summaries exclude three collapsed
encoders, two shared and one constant. We set this exclusion after probing: these
encoders' between-image distance had collapsed to zero, and they scored zero.
Including them changes the shared \byol{} median from 0.155 to 0.144 and the
constant median from 0.145 to 0.144.

\mae{} was trained on the single-view images with the same optimiser, but
used a different mask scheme and decoder, with pixels as prediction targets.

\subsection{Probe and Movement Measurement}

The readability probe has two cross-attention blocks of width 64 with four
heads, position-only queries built from 22 Fourier
features, Fourier token positions, and a learned locality prior with
$\sigma=0.25$. A layer norm and linear output map predict the standardised
targets. The joint probe has \probejointparams{} parameters. The
single-target probe differs only in its output head and selects its
checkpoint using $z_1$ alone. The small probe has \probesmallparams{} parameters: one
single-head cross-attention step using six position features, with no
locality prior, and a linear output for the same three targets.

For each component, let $\mu_\alpha(E)$ and $\sigma_\alpha(E)$ be the
simulator's mean and standard deviation of $\ln\alpha$ at energy $E$,
with $\mu_T(E)$ and $\sigma_T(E)$ defined for $\ln T$ in the same way.
Let $\rho(E)$ be their correlation. The shape targets are
\[
\begin{aligned}
 u=\frac{\ln\alpha-\mu_\alpha(E)}{\sigma_\alpha(E)},\qquad & v=\frac{\ln T-\mu_T(E)}{\sigma_T(E)},\\[4pt]
 z_1=\frac{v+u}{\sqrt{2(1+\rho(E))}},\qquad & z_2=\frac{v-u}{\sqrt{2(1-\rho(E))}}.
\end{aligned}
\]
The simulator draws $\ln\alpha$ and $\ln T$ from two independent standard
normal variables per component, and $z_1$ and $z_2$ recover them exactly.
A fit of $z_1$ from energy alone scored at or below zero.
The energy target uses the
mean and standard deviation of $\ln E$ in the probe-fitting set; the same
values are applied to the selection and test sets.
Across settings, the median readability scores of $z_1$ and $z_2$ have a correlation of \shapereadoutcorr{}, so the main text reports $z_1$.

Probes train with Adam, a learning rate of $2\times10^{-3}$, batch size
128, and mean-squared error against the standardised targets. The fitting,
selection and test sets are generated separately, with
\probeFittingEvents{}, \probeHeldoutEvents{} and \probeTestEvents{} images.
Every 100 steps the selection set is scored, and the checkpoint with the
lowest median absolute error, averaged over the targets, is selected. The test
set is scored once, at that checkpoint. On the test set the baseline's median
absolute error on $z_1$ is \constantmedaezone{}; $d(z_1)$ is the part of that error
the probe removes.

Movement is $r = d_{\mathrm{intra}}/d_{\mathrm{inter}}$, where
$d_{\mathrm{intra}}$ is the mean cosine distance between replicas of the
same image and $d_{\mathrm{inter}}$ the mean distance between different
images. Movement evaluations use 64 images with eight replicas each; the
between-image distance uses the first replica of each image. The reported
per-token movement is the median of the ratios over token positions.

\subsection{Probe Controls}

\textbf{Positive control.} \mae{} shows that the probe can read the factor
from a trained encoder. Its median score was 0.283, and at the end of
training every \mae{} encoder scored above every other trained encoder. Single-view \ijepa{}, trained on the
same images, scored \singleviewmasked{}, lower than in the shared and constant conditions.
The single-view and two-view constructions differ both in how the images
are made and in the available context, and we did not isolate which difference matters.

\textbf{Encoder copy.} Target and online copies were compared in 24 pairs:
twelve \ijepa{} encoders, six shared and six constant, with two probe fits on
the \probebig{} probe. The largest readability difference was
\readoutboundcontent{}, against gaps of about
\readoutobjectivegap{} between \ijepa{} and \simclr{}. The largest per-token movement difference was
\readoutboundmovement{}, about half the lowest \simclr{} movement. With either copy, five of the six shared
encoders scored below every shared or constant \simclr{} encoder.

\textbf{Joint and single-target prediction.} Joint prediction raised scores
in every tested condition, including the untrained encoder's. For single-view
\ijepa{}, the increase accounted for almost all of the joint median. On every
probe, shared \ijepa{} with the default predictor had a lower median than
shared \simclr{}, and both remained below \mae{}.

\begin{table}[t]
\centering
\caption{Readability of $z_1$ in the shared condition on four probes at
\budget{} steps: medians over training seeds with probe fits averaged,
\byol{} without its collapsed encoders. The standardised probe is the joint
probe with fixed standardisation in place of its token layer norm. On the
first three probes, the lowest-scoring
encoder with the wider predictor scored above the highest \simclr{} or
\byol{} encoder: 0.222 against 0.193 on the joint probe, 0.209 against 0.157
on the single-target probe and 0.148 against 0.109 on the 7.8k probe. On the
standardised probe, the highest \simclr{} encoder scored 0.244 and the lowest
wider one 0.231. The 7.8k-parameter probe's
untrained fits were still improving at the budget, so that untrained score
is a lower bound.}
\label{tab:instruments}
\small
\begin{tabular}{lrrrrrr}
\toprule
 & \multicolumn{3}{c}{\ijepa{} predictor} & & & \\
\cmidrule(lr){2-4}
Probe & default & deeper & wider & \simclr{} & \byol{} & untrained \\
\midrule
\probebig{} joint         & 0.061 & 0.184 & 0.233 & 0.153 & 0.155 & 0.310 \\
\probebig{} single-target & 0.028 & 0.166 & 0.216 & 0.121 & 0.128 & 0.256 \\
7.8k joint                & 0.020 & 0.118 & 0.154 & 0.078 & 0.079 & 0.074 \\
\probebig{} joint, standardised & 0.060 & 0.185 & 0.233 & 0.204 & 0.153 & 0.338 \\
\bottomrule
\end{tabular}
\end{table}

\textbf{Probe budget.} The untrained score is a median over \referenceLongInits{}
random initialisations. For the final trained encoders, the
main-probe fits reached their best selection-set evaluation within
\trainedbeststepmax{} steps; the untrained fits kept improving until between \untrainedbeststeprange{} steps.
None of the trained medians moved between
10{,}000 and \budget{} steps. The untrained score therefore depends on the budget.

\textbf{Raw input.} The probe was also fitted to the input image cut into the
encoder's \trainingPatchSpec{} patches, with no encoder, at the same budget. It scored \rawinputref{} for $z_1$, inside the range of the
\referenceLongInits{} untrained initialisations (0.285 to 0.325). Replacing the
probe's per-token layer norm with fixed per-dimension standardisation raised
the raw-patch score to 0.396. Applied to the encoders in Table~\ref{tab:instruments}, the same change raised the untrained and \simclr{} scores and moved the \ijepa{} and \byol{} medians by at most 0.002. The highest of these encoders then scored 0.244, and \mae{}, the trained encoder closest to untrained on the main probe, at most 0.287. Both stayed below the lowest untrained initialisation, 0.328.

\section{Additional Comparisons}
\label{app:seeds}

\subsection{Predictor Capacity}

\begin{table}[t]
\centering
\caption{Single-view predictor sweeps trained on mixed images, using the main
joint probe at \budget{} steps. Entries are medians over seeds; an untrained
encoder's median is \untrainedrefforty{}. Each width has six seeds, and width 48
used different training seeds from the other widths. Depths 1 and 6 have
three seeds each; the depth-6 median rests on two seeds near it and one
below. The width-48, depth-3 setting is the mixed single-view curve in
Figure~\ref{fig:dietcurves}.}
\label{tab:singleviewcapacity}
\setlength{\tabcolsep}{4pt}%
\begin{tabular}{lccccc}
\toprule
Width (depth 3) & 12 & 24 & 48 & 96 & 192 \\
\midrule
$d(z_1)$ & 0.037 & \widthcontentmin{} & 0.106 & \widthcontentmax{} & 0.220 \\
\bottomrule
\end{tabular}
\hspace{1.5em}
\begin{tabular}{lccc}
\toprule
Depth (width 48) & 1 & 3 & 6 \\
\midrule
$d(z_1)$ & 0.058 & 0.106 & 0.221 \\
\bottomrule
\end{tabular}
\end{table}

In the single-view sweeps on mixed images, readability rose from width 24 to
the encoder's width, with no gain beyond it, and rose with depth
(Table~\ref{tab:singleviewcapacity}).

In the shared condition, two of the three deeper-predictor seeds that reached the wider predictor's level were also
among the three highest under the default predictor.

We probed shared \ijepa{} under the three predictors over training to epoch
500 (Figure~\ref{fig:trajectory}). Under the default predictor, readability
held to epoch 150 and then declined. Under the deeper and the wider predictor
there was no sustained late decline. Shared \simclr{} moved the other way: it
was below \ijepa{} with the default predictor at every seed from epoch 20 to
100, and above it at five of six seeds at epochs 400 and 500.

\begin{figure}[t]
  \centering
  \includegraphics[width=3.71in,trim=0 7.8bp 0 8.4bp,clip]{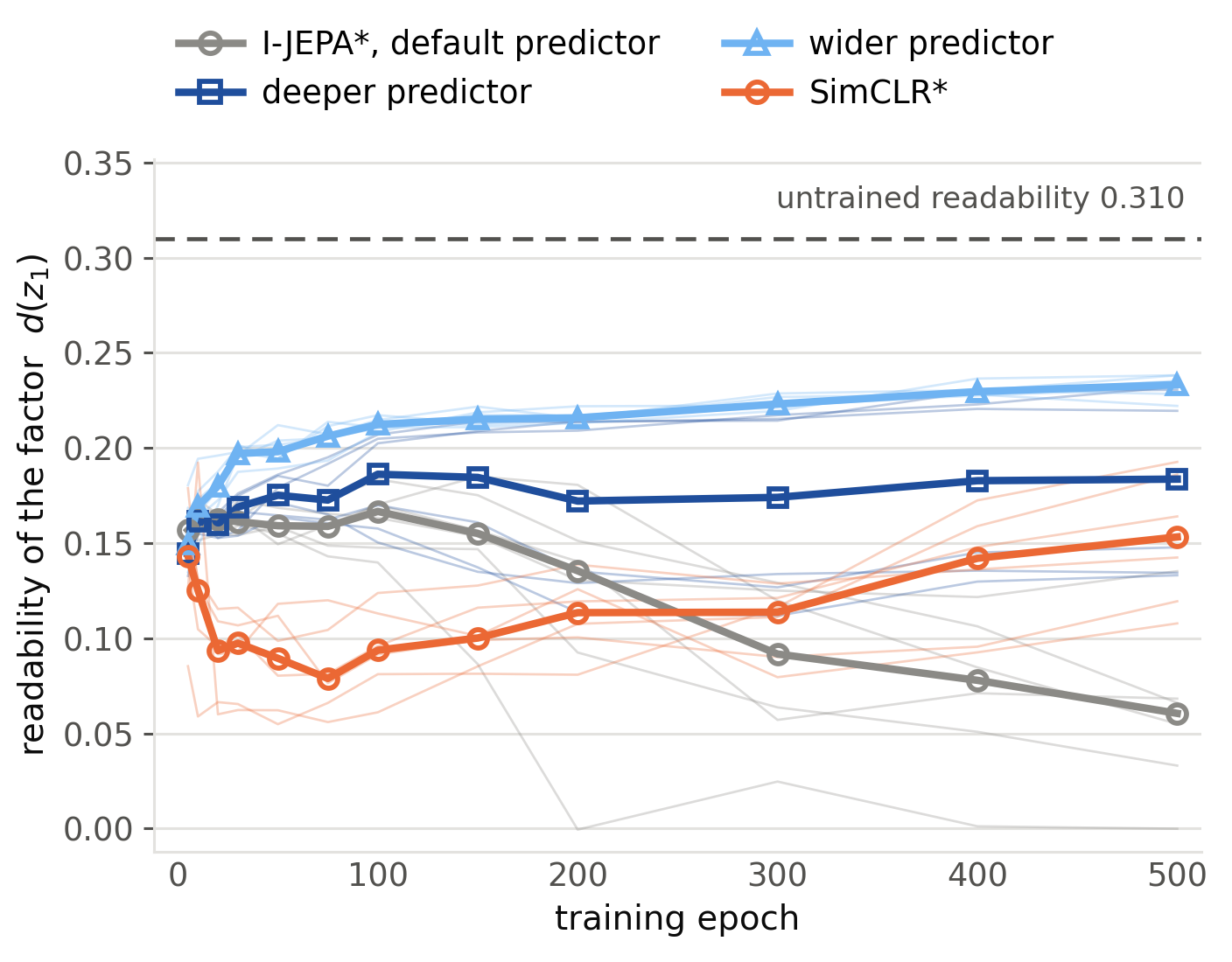}
  \caption{Readability of the shared factor over training, for shared \ijepa{}
  under the three predictors and for shared \simclr{}. Thin lines are training
  seeds and thick lines medians. The dashed line is the untrained encoder.}
  \label{fig:trajectory}
\end{figure}

\subsection{Energy}\label{app:energy}

Shared \ijepa{} had a median energy score $d(E)$ of 0.314, against 0.389 for shared \simclr{} and 0.494
for the untrained encoder. In the constant condition, \ijepa{} was also below
\simclr{}. With the deeper and the wider predictor, the energy
score of shared \ijepa{} rose to 0.413 and 0.457, so energy, like shape, moved
toward the untrained level as predictor capacity grew.

\subsection{Training Graininess}\label{app:diet}

The untrained score increased as images became smoother (Figure~\ref{fig:dietcurves}), yet the grainiest
training images produced the highest trained scores in both conditions. With
smoother probe images, readability rose for encoders trained on grainy and mixed
images but changed little for those trained on default and smooth images.

\begin{figure}[t]
  \centering
  \includegraphics[width=\textwidth,trim=0 13.0bp 0 8.8bp,clip]{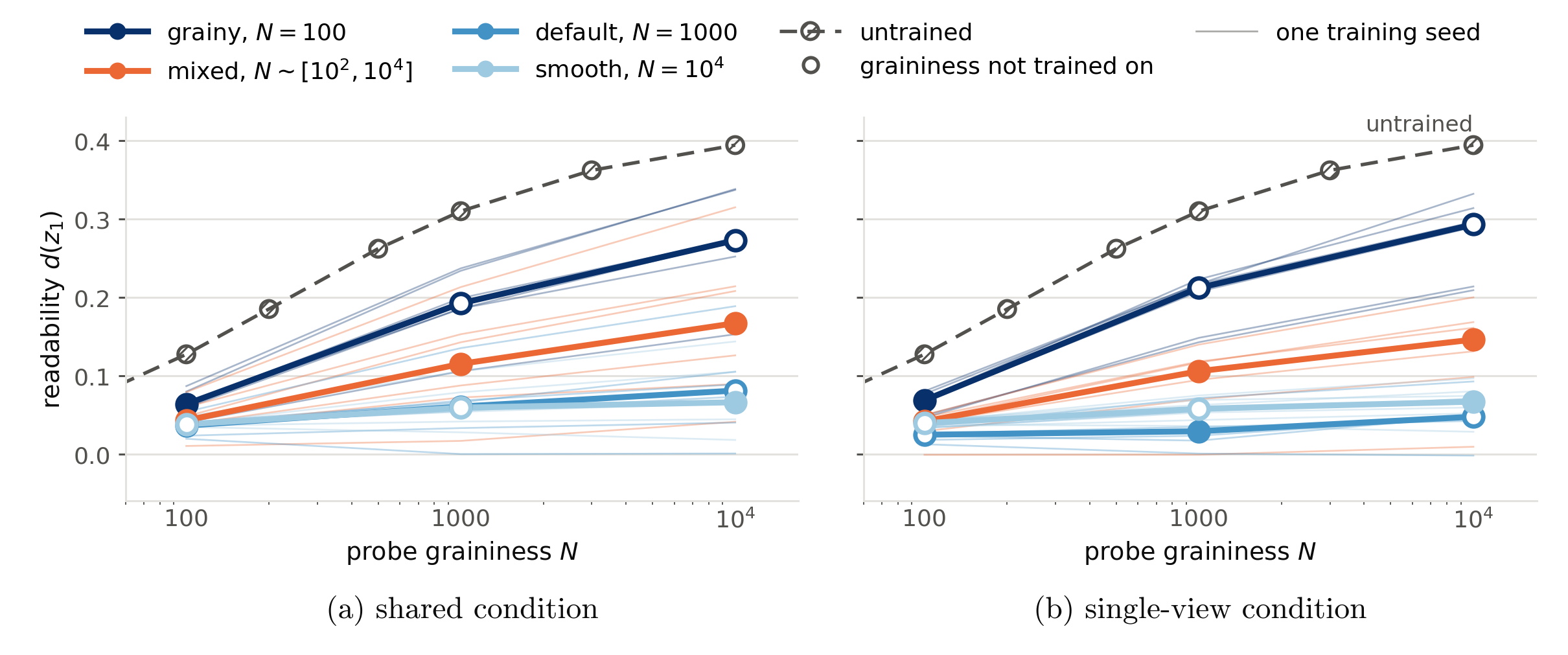}
  \caption{Readability against probe graininess for the four choices of training graininess,
  in the shared (a) and single-view (b) conditions, at \budget{} steps.
  Thin lines are training seeds, thick lines
  medians; a filled marker is a graininess the encoders trained on, a hollow
  one a graininess they did not. The dashed curve is the untrained
  encoder over seven graininesses.}
  \label{fig:dietcurves}
\end{figure}

In the shared condition, we also dropped the grainiest images from the mixed
set, so that $N$ starts at 200 instead of 100. Encoders trained this way
scored a median of \floorlevelmed{} at the default probe graininess, between the mixed- and default-trained
encoders. The change also made the training images easier to read: averaged
over the training graininesses, the untrained score rose from below the
default-image score to slightly above it. So the test changed two things at once, how grainy
the training images were and how easy they were to read, and cannot tell which
of the two lowered the score.

\end{document}